\documentclass[10pt,twocolumn,letterpaper]{article}

\usepackage[pagenumbers]{cvpr} % To force page numbers, e.g. for an arXiv version

\usepackage{graphicx}
\usepackage{amsmath}
\usepackage{amssymb}
\usepackage{booktabs}
\usepackage{multirow}
\usepackage[pagebackref,breaklinks,colorlinks]{hyperref}

\usepackage[capitalize]{cleveref}
\crefname{section}{Sec.}{Secs.}
\Crefname{section}{Section}{Sections}
\Crefname{table}{Table}{Tables}
\crefname{table}{Tab.}{Tabs.}

\def\cvprPaperID{*****} % *** Enter the CVPR Paper ID here
\def\confName{CVPR}
\def\confYear{2022}

\begin{document}

% \begin{sloppypar}
\title{An Adaptive Spatial-Temporal Local Feature Difference Method \\ for Infrared Small-moving Target Detection}

\author{$^{1}$Yongkang Zhao, $^{1}$Chuang Zhu$^{\ast}$, $^{2}$Yuan Li, $^{2}$Shuaishuai Wang , $^{1}$Zihan Lan, $^{1}$Yuanyuan Qiao\\
$^{1}$School of Artificial Intelligence, Beijing University of Posts and Telecommunications.\\
$^{2}$School of Computer Science, Peking University.\\
{\tt\small czhu@bupt.edu.cn}}

% \author{Yongkang Zhao\\
% School of Artificial Intelligence, Beijing University of Posts and Telecommunications.\\
% Beijing 100876, China.\\
% \and
\maketitle

%%%%%%%%% ABSTRACT
\begin{abstract}
Detecting small moving targets accurately in infrared (IR) image sequences is a significant challenge. To address this problem, we propose a novel method called spatial-temporal local feature difference (STLFD) with adaptive background suppression (ABS). Our approach utilizes filters in the spatial and temporal domains and performs pixel-level ABS on the output to enhance the contrast between the target and the background. The proposed method comprises three steps. First, we obtain three temporal frame images based on the current frame image and extract two feature maps using the designed spatial domain and temporal domain filters. Next, we fuse the information of the spatial domain and temporal domain to produce the spatial-temporal feature maps and suppress noise using our pixel-level ABS module. Finally, we obtain the segmented binary map by applying a threshold. Our experimental results demonstrate that the proposed method outperforms existing state-of-the-art methods for infrared small-moving target detection.

\end{abstract}

%%%%%%%%% BODY TEXT
\section{Introduction}

%-------------------------------------------------------------------------
\begin{figure*}[!htbp]
\begin{center}
\includegraphics[width=0.99\textwidth]{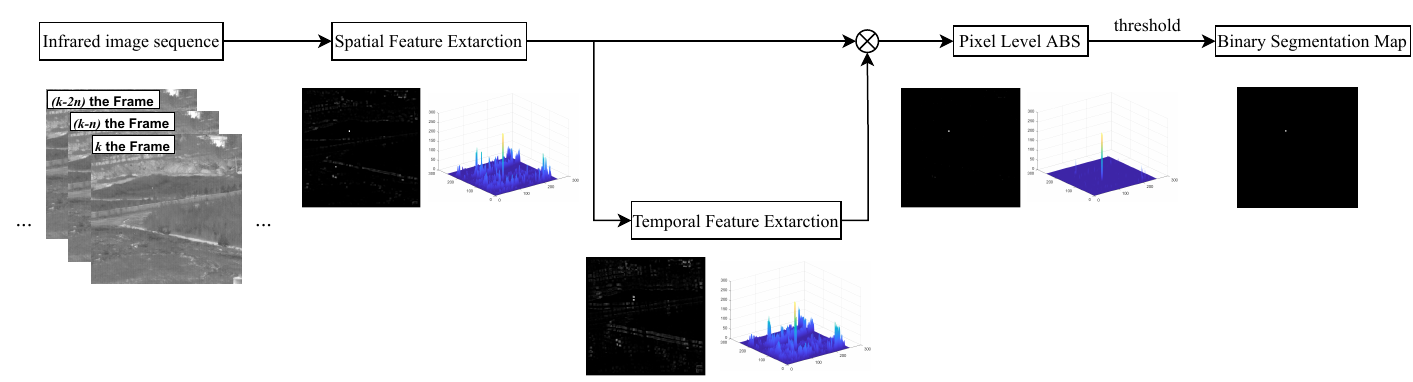}
\end{center}
   \caption{The framework of the proposed STLFD method for IR small moving target detection. The detection process mainly includes spatial feature extraction of the \textit{k}th frame, temporal feature extraction of \textit{k}th, \textit{(k-n)}th and \textit{(k-2n)}th frames, and pixel-level ABS module to calculate the output of the spatial and temporal domain extractor.}
\end{figure*}

The detection and tracking of small moving targets in infrared (IR) image sequences play a vital role in multiple fields, including aerospace, unmanned aerial vehicle (UAV) shooting, night shooting, and guidance \cite{detect_move_survey}. However, accurately identifying small targets in IR image sequences presents a significant challenge due to their low energy intensity, lack of distinct semantic texture features, and potential confusion with background noise \cite{STTM-SFR}. Furthermore, the use of drones for photography introduces additional complexities such as background movement, further complicating the detection of small moving targets in diverse and intricate scenes. Consequently, the detection of small moving targets in various complex IR image sequences remains a formidable task.

In recent years, several studies have focused on utilizing deep learning methods for IR small target detection \cite{deep_learning1, deep_learning2}. However, these methods face issues such as a lack of sufficient public data to support effective training, and the high complexity of models leading to the inability to meet the demand for real-time and lightweight requirements in real applications. Therefore, we only discuss methods without deep learning in this paper. Current studies mainly rely on two branches: the single-frame spatial filtering methods \cite{single_frame_survey} and the combination of multi-frame spatial and temporal filtering methods \cite{frame_difference,background_subtraction}.

Most single-frame spatial domain filtering methods are mainly based on the local contrast information in space to enhance the target region and suppress the background region. Chen \textit{et al.} \cite{LCM} proposed the local contrast method (LCM), which first used a patch to calculate the difference between the target and the surrounding area, enhancing the target area. Building on the LCM, Han \textit{et al.} \cite{RLCM} proposed the multi-scale relative local contrast measure (RLCM). Wei \textit{et al.} \cite{MPCM} proposed the multi-scale patch-based contrast measure (MPCM), which combined gradients in each direction to enhance targets with various energy intensities. Wu \textit{et al.} \cite{DNGM} introduced the double-neighborhood gradient method, which defined a tri-layer window to detect multi-scale targets. However, single-frame based methods that rely solely on spatial domain information may not effectively suppress background noise. Additionally, these methods are limited by the use of only single-frame 2-D features, which may not distinguish the target from some backgrounds with disturbing properties. As small targets in IR images have motion characteristics, methods that combine spatial and temporal information have been proposed to address these limitations.

In recent years, several methods combining multi-frame spatial and temporal filtering have made significant progress. Deng \textit{et al.} \cite{STLCF} proposed the spatial–temporal local contrast filter (STLCF) method based on local contrast filter, which used a mean filter to calculate the feature map on both spatial and temporal domains. Du and Hamdulla \cite{STLDM} proposed the spatial–temporal local difference measure (STLDM) method, which directly calculated the IR image sequence as a 3-D spatial-temporal domain. Pang \textit{et al.} \cite{NSTSM} proposed the spatiotemporal saliency method (NSTSM) for low-altitude slow small target detection, which used a tri-layer window to calculate the mean of the variance difference between the internal and external windows. Zhang \textit{et al.} \cite{STVDM} proposed the spatial-temporal local vector difference (STVDM) measure, which converted spatial information into vectors by calculating the cosine similarity to obtain the target information. Although the methods mentioned above incorporate temporal domain information, there are still two key issues to consider. Firstly, previous methods use future information in the video sequence, which generates an unavoidable delay that varies with the amount of future information used and can not meet the need for real-time in some scenarios. Secondly, camera movement can cause the background to be perceived as motion information, while the previous method directly uses the temporal domain information between frames, which cannot effectively avoid such problems.

To address the challenges mentioned above, we propose a novel approach for small target detection in IR image sequences, called the spatial-temporal local feature difference (STLFD) method. Our method focuses on extracting spatial gray value features in IR image sequences using a specially designed spatial domain filter. By utilizing the spatial feature of the current frame along with previous frames as temporal domain information, we eliminate the need for future information and mitigate output delays. Additionally, we introduce a pixel-level adaptive background suppression (ABS) module to enhance the contrast of the output feature maps and effectively suppress noise generated by background motion. The implementation details of our method are described in the following sections. We demonstrate the effectiveness of our approach through experimental results.

\section{Proposed Method}

The detection framework of the algorithm is illustrated in Fig. 1. Here, given an IR image sequence, we only use the information of the current frame and its previous frames, we choose the \textit{k}th frame, the\textit{ (k - n)}th frame and the \textit{(k - 2n)}th frame as input. The input is then passed through a spatial filter to generate the spatial feature map (Smap). Temporal feature information will be extracted to obtain the temporal feature map (Tmap) based on Smap. To obtain the spatial-temporal feature map (STmap) of the \textit{k}th frame, the Smap and Tmap are element-wise multiplied. Finally, a pixel-level ABS is applied to enhance the target and suppress the background, resulting in the final output feature map. The binary segmentation results are obtained by applying a threshold to the final output feature map.

\begin{figure}[!h]
\center
\includegraphics[scale=0.6]{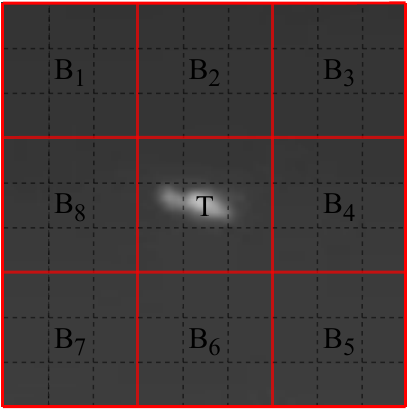}
   \caption{Structure of the spatial filter kernel in one frame.}
\end{figure}
% \noindent
\subsection{Spatial Feature Extraction} 

\begin{table*}[!htbp]
\caption{The detailed description of five real IR image sequences}
\centering
% \resizebox{\textwidth}{!}{
\begin{tabular}{ccclc}
\toprule[1.5pt]
\multicolumn{1}{l}{} & \multicolumn{1}{l}{Frame Number} & \multicolumn{1}{l}{Image Scale} & Scene Description                                                                                                                                          & \multicolumn{1}{l}{Target Type} \\ \midrule[1pt]
Seq. 1                 & 745                              & 256$\times$256                         & \begin{tabular}[c]{@{}l@{}}Ground-sky background\\ Background movement and jitter\\ Strong road and mountain noise impact\end{tabular}                     & UAV                             \\\midrule[1pt]
Seq. 2                 & 1500                             & 256$\times$256                         & \begin{tabular}[c]{@{}l@{}}Ground-sky background\\ Background slow movement and jitter\\ Strong ground noise impact\end{tabular}                           & UAV                             \\\midrule[1pt]
Seq. 3                 & 750                              & 256$\times$256                         & \begin{tabular}[c]{@{}l@{}}Low-altitude background\\ Background fast movement and large jitter\\ Strong building noise impact\end{tabular}                 & UAV                             \\\midrule[1pt]
Seq. 4                 & 500                              & 256$\times$256                         & \begin{tabular}[c]{@{}l@{}}Ground-sky background\\ Background slow movement and slight jitter\\ Strong woods noise and edge background impact\end{tabular} & UAV                             \\ \midrule[1pt]
Seq. 5                 & 100                              & 320$\times$240                         & \begin{tabular}[c]{@{}l@{}}sky bacground\\ Static bacground\\ Cloud noise impact and low target contrast\end{tabular}                                      & UAV                             \\ \bottomrule[1.5pt]
\end{tabular}
% }
\end{table*}

The spatial filter kernel is shown in Fig. 2. We define a 3$\times$3 area as a patch \cite{LCM}, and a filter kernel is formed by 9 adjacent patches. The patch in the middle area is the target patch $T$, while the patch in the surrounding area is the background patch $B_{i}\ (i = 1, 2, ..., 8)$. The purpose of our spatial filter is to obtain the local contrast information between the target patch $T$ and the surrounding background patch $B_{i}$ of the \textit{k}th frame, so as to suppress the background and enhance the target features. To be able to obtain the local contrast information, we calculate the difference between the target patch and the surrounding background patch separately, and the function is defined as follows:
\begin{small}
\begin{align}
D_{n} = max\{2 \cdot MAX_{T} - MEAN_{B_{n}} - MEAN_{B_{n+4}}, 0\},\nonumber \\ (n = 1, 2, 3, 4),
\end{align}
\end{small}
% \begin{gather}
% D_{n} = 2 \cdot MAX_{T} - MEAN_{B_{n}} - MEAN_{B_{n+4}}, (n = 1, 2, 3, 4),\\
% D_{n} = max\{D_{n}, 0\}, (n = 1, 2, 3, 4),
% \end{gather}
where $MAX_{T}$ and $MEAN_{B_{n}}$ denote the maximum value of patch $T$ and the mean value of patch $B_{n}$, respectively. Therefore, we can obtain four local gradient contrast values ${D_{1},\ D_{2},\ D_{3},\ D_{4}}$ to characterize the information of the current area. For patch $T$, we use maximum filtering to effectively enhance the energy intensity of weak targets and their area on the output feature map.

To obtain the final spatial feature map, we calculate the product of the maximum and minimum of the four contrast values $D_{1},\ D_{2},\ D_{3},\ D_{4}$, then normalize the result. This process results in a spatial feature map named $Smap$, which is given by the following function:

\begin{gather}
Smap(i, j) = max(D_{n}) \cdot min(D_{n}), (n = 1, 2, 3, 4), \\
Smap(i, j) = \frac{Smap(i, j)}{max_{i, j}\{Smap(i, j)\}}.
\end{gather}

The output of $Smap$ is a normalized representation of the local gradient contrast values that are indicative of the target present in the current region.

It is worth noting that the calculation of $D_{n}$ only retains values greater than zero since including negative values can affect the calculation of the area, where the central gray value is lower than the surrounding area. By excluding negative values, we ensure that $Smap$ reflects the true local gradient values that are indicative of the target present in the current area. Furthermore, by calculating the product of the maximum and minimum non-negative gradient values, we can effectively suppress false detection regions caused by factors such as edges. This is beneficial for subsequent spatial-temporal feature multiplication and ABS operations, as it reduces false positives and improves the accuracy of the target detection system.

\subsection{Temporal Feature Extraction} 

Although the spatial filter is effective in enhancing target regions, it has limited ability to suppress non-target regions such as temporal noise and edge noise. To address this issue, we introduce a temporal domain filter. Unlike conventional temporal domain filtering for images, we perform filtering in time based on the target-enhanced features obtained from the spatial filter. To account for the real-time factor and limited memory constraints on some hardware, we only use information from the previous frames for feature calculation. Specifically, we calculate the maximum and minimum temporal domain values for the current $k$th frame, $(k-n)$th frame, and $(k-2n)$th frame, respectively, to extract the motion vector of the target. The function is defined as follows:

\begin{footnotesize}
\begin{gather}
T_{max} = max\{Smap_{k_{i, j}}, Smap_{(k-n)_{i, j}}, Smap_{(k-2n)_{i, j}}\}, \\
T_{min} = min\{Smap_{k_{i, j}}, Smap_{(k-n)_{i, j}}, Smap_{(k-2n)_{i, j}}\},
\end{gather}
\end{footnotesize}
where $Smap_{k_{i, j}}$ denotes the spatial feature value of the pixel at the $(i, j)$ coordinate of the $k$th frame. The final  $Tmap$ is defined as follows:
\begin{gather}
Tmap = T_{max} - T_{min}, \\
Tmap(i, j) = \frac{Tmap(i, j)}{max_{i, j}\{Tmap(i, j)\}}.
\end{gather}

\subsection{Pixel Level ABS} 

The detection feature obtained from the spatial filter and the temporal filter is defined as:
\begin{align}
STmap(i, j) = Smap(i, j) \cdot Tmap(i, j).
\end{align}
Although this map has been able to get more accurate features, it still cannot suppress some edges, especially for shooting camera movement. Therefore, we designed the pixel-level ABS module to further suppress non-target areas and improve the contrast between the target area and the background area. We define an area of p$\times$p pixels as an ABS kernel and the final STLFD map is defined as follows:
\begin{align}
STLFD = \begin{cases}
I_{i, j}, & I_{i, j} = ABS_{i,j} \\
I_{i, j} \cdot ABS_{i,j}, & else \\
\end{cases},
\end{align}
where $I_{i, j}$ denotes the feature map value at $(i, j)$ of the $STmap$, and $ABS_{i,j}$ denotes the maximum value within an ABS kernel size region centered at $(i, j)$ of the $STmap$. Since the value of the $STmap$ ranges from 0 to 1, multiplying the maximum value by the non-maximum value in the ABS kernel can suppress the surrounding area based on the feature value of that area.

\subsection{Threshold Segmentation Binary Map} 

After obtaining the output feature map of STLFD, we use the dynamic threshold to obtain the segmented binary output map based on the general method of selecting the threshold. The threshold is defined as follows:
\begin{align}
Threshold = \mu_{STLFD} + k \times \sigma_{STLFD},    
\end{align}
where $\mu_{STLFD}$ and $\sigma_{STLFD}$ denote the mean value and standard value of the STLFD map, respectively, and $k$  is a hyper-parameter that needs to be adjusted empirically.

% 实际输出图和三维图
\begin{figure}[!ht]
\begin{center}
\includegraphics[width=0.45\textwidth]{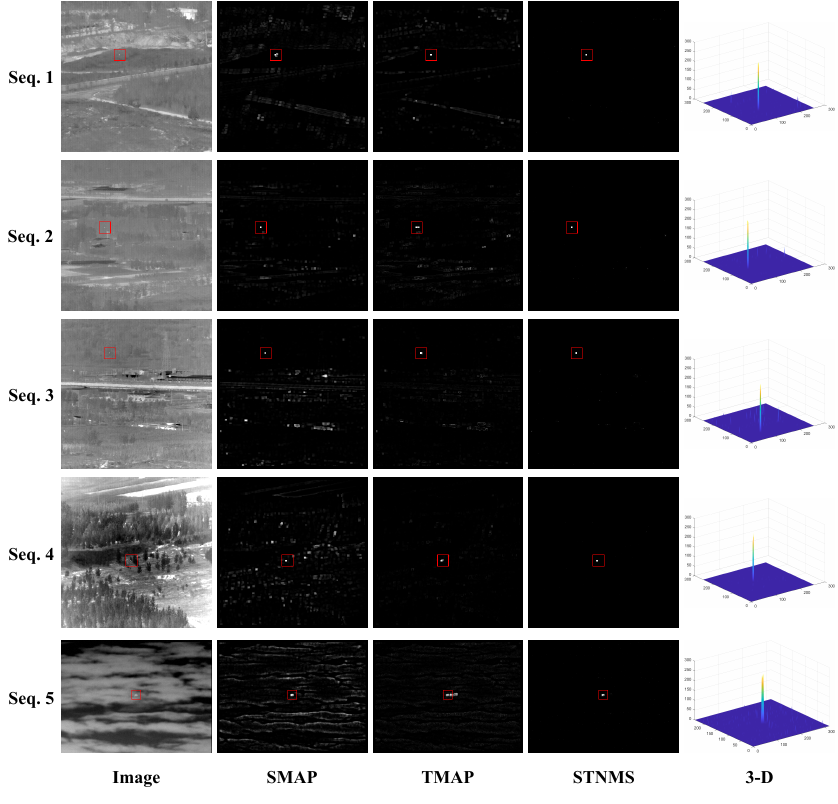}
\end{center}
   \caption{The detection results of the proposed method on five IR image sequences. where the red rectangular box identifies the position of the groundtruth target. The results, from left to right are input image, spatial map, temporal map, spatial-temporal map with ABS module, and 3-D output feature map.}
\end{figure}

% 方法对比图
\begin{figure*}[!ht]
\begin{center}
\includegraphics[width=0.99\textwidth]{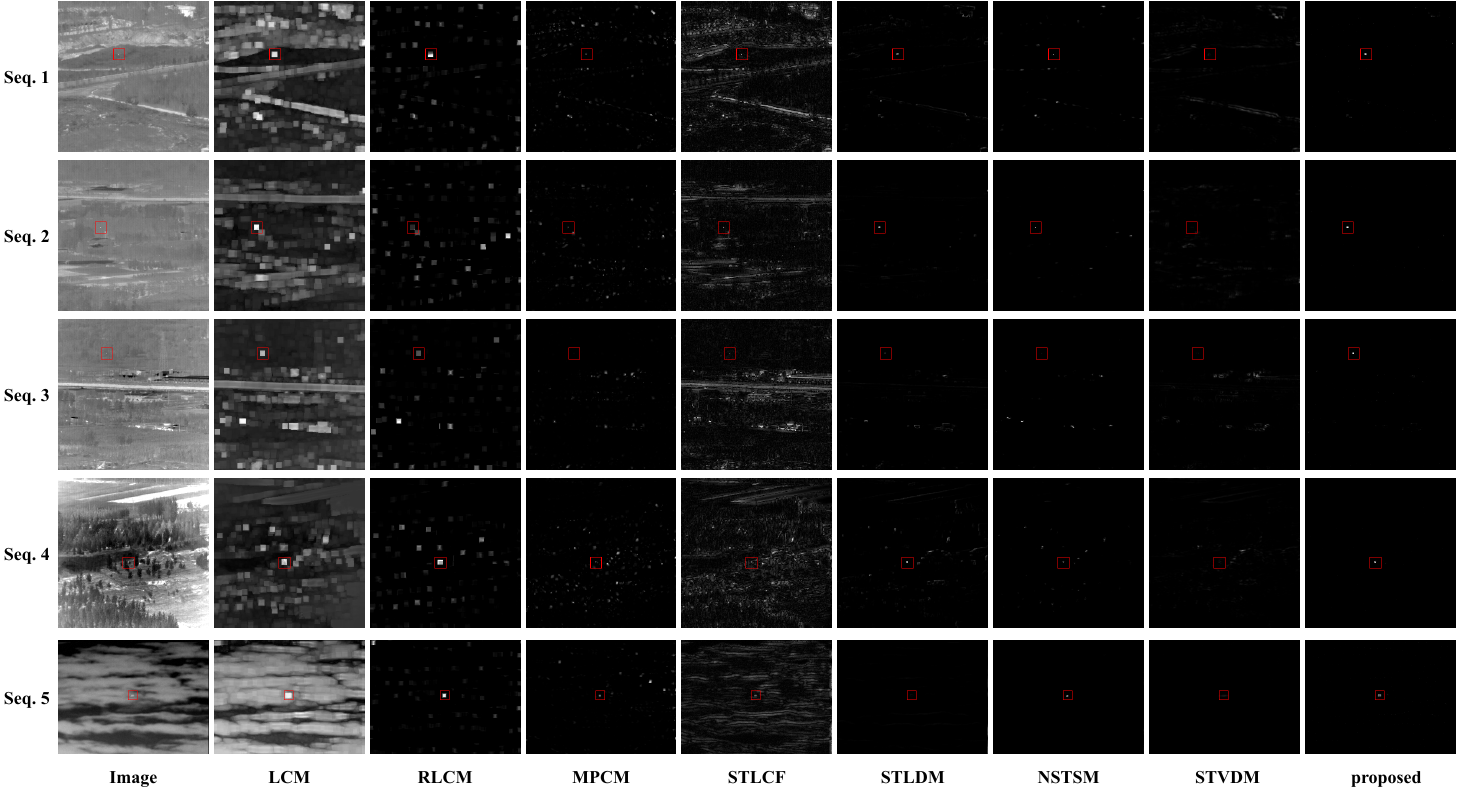}
\end{center}
   \caption{Quality comparison of the detection results of the compared methods with the proposed method on the five IR image sequences. The red rectangular box identifies the position of the groundtruth target. The results from left to right are input image, LCM, RLCM, MPCM, STLCF, STLDM, NSTSM, STVDM, and proposed STLFD}
\end{figure*}

\section{Experiment}

We conducted experiments to assess the effectiveness of the proposed method, by comparing it against multiple baseline methods on five IR image sequences. For all experiments, we used fixed hyper-parameter settings, with an ABS kernel width of 15$\times$15 and a time gap of 5, which remained consistent across all tests.

\subsection{Datasets and Baseline Methods}

Table 1 provides a detailed description of the IR image sequences used in the experiment. Seq. 1-4 \cite{data_22scene} include shooting camera movement and jitter, which increase the difficulty of detecting weak targets in motion. Seq. 5 features a static cloud background with low contrast between the background and the target. To evaluate the performance of the proposed method, we compared it against several classical single-frame detection methods and multi-frame detection methods. The single-frame detection methods used in the comparison were LCM \cite{LCM}, RLCM \cite{RLCM}, and MPCM \cite{MPCM}. The multi-frame detection methods included STLCF \cite{STLCF}, STLDM \cite{STLDM}, NSTSM \cite{NSTSM}, and STVDM \cite{STVDM}. The parameter settings for all methods were consistent with those described in the original paper.

% SCRG和BSF对比表
\begin{table*}[!htbp]
\caption{Average SCRG and BSF obtained from each method on the five IR image sequences}
\resizebox{\textwidth}{!}{
\begin{tabular}{clllllllll}
\toprule[1.5pt]
\multicolumn{1}{l}{}  & Seq. & LCM     & RLCM    & MPCM    & STLCF   & STLDM            & NSTSM            & STVDM   & Proposed         \\ \midrule[1pt]
\multirow{5}{*}{SCRG} & 1    & 15.1309 & 15.5328 & 7.6966  & 6.8236  & 73.6724          & 72.3106          & 18.5422 & \textbf{82.4716} \\
                      & 2    & 13.9867 & 15.0874 & 10.3713 & 3.0633  & 80.8504          & 81.8743          & 13.9208 & \textbf{91.1291} \\
                      & 3    & 14.2249 & 14.5249 & 10.1730 & 4.9626  & 91.0629          & 82.8706          & 69.5515 & \textbf{94.3947} \\
                      & 4    & 19.2291 & 18.1442 & 14.9650 & 11.0055 & \textbf{73.5611} & 22.2123          & 17.9511 & 27.1421          \\
                      & 5    & 3.2595  & 9.9861  & 11.8585 & 1.4126  & 8.9046  & \textbf{88.8715} & 5.9907  & 14.1818          \\ \midrule[1pt]
\multirow{5}{*}{BSF}  & 1    & -3.2808 & -0.1328 & 6.4999  & 0.0001  & 60.8000          & 76.6493          & 10.2905 & \textbf{78.7617} \\
                      & 2    & -3.2215 & -0.4893 & 6.9551  & -0.9537 & 73.6884          & 78.6917          & 9.1504  & \textbf{82.4908} \\
                      & 3    & -4.4437 & -2.0283 & 11.6067 & 0.1224  & 83.3581          & 84.4983          & 78.2508 & \textbf{83.9236} \\
                      & 4    & -0.1651 & 1.8001  & 6.5400  & 2.3709  & \textbf{65.6871} & 18.9929          & 13.1851 & 16.7844          \\
                      & 5    & -0.8221 & 3.4637  & 13.4629 & 4.1208  & 20.6295 & \textbf{91.3577} & 13.1702 & 12.0608          \\ \bottomrule[1.5pt]
\end{tabular}
}
\end{table*}

% ROC曲线图
\begin{figure*}[!ht]
\begin{center}
\includegraphics[width=0.99\textwidth]{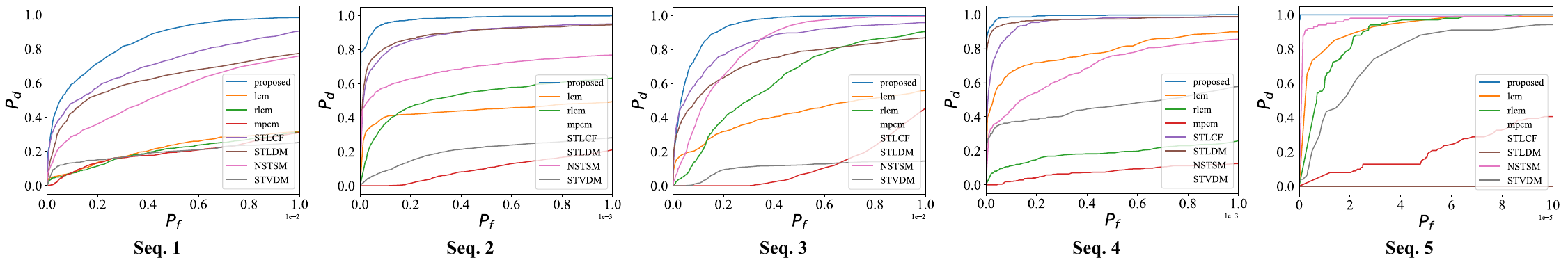}
\end{center}
   \caption{ROC curve of the detection result of all the methods on five real IR image squences.}
\end{figure*}

\begin{figure*}[!ht]
\begin{center}
\includegraphics[width=0.99\textwidth]{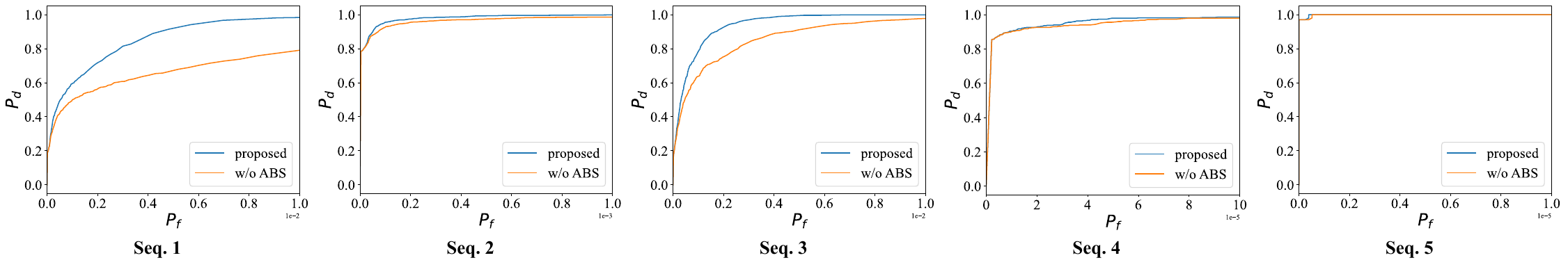}
\end{center}
   \caption{ROC curves for ablation study of the proposed method with and without the pixel-level ABS module.}
\end{figure*}

\subsection{Evaluation Metrics}

There are several metrics used to evaluate the performance of IR small target detection algorithms. The main metric is the receiver operating characteristic (ROC) curve, which directly measures the detection performance by analyzing the relationship between the probability of detection ($P_{d}$) and the probability of false alarm ($P_{f}$)  \cite{ROC, STLDM}. In addition to the ROC curve, there are other metrics that indirectly evaluate the performance of small target detection algorithms. These include the signal-to-clutter ratio gain (SCRG) and the background suppression factor (BSF). These two metrics reflect the local contrast enhancement effect of the target area and the suppression effect of background noise, which are used to measure whether a target is relatively easy to detect. We take the ROC curve as our main evaluation metric, where $P_{d}$ and $P_{f}$ are defined as: 
\begin{gather}
P_{d} = \frac{n_{t}}{N_{t}}, \\
P_{f} = \frac{n_{f}}{N},
\end{gather}
where $n_{t}$, $N_{t}$, $n_{f}$, $N$ denote the number of detected true targets, the total number of real targets, the number of pixels of detected false targets and the total number of pixels in the image sequences, respectively. We obtain the relationship curve between $P_{d}$ and $P_{f}$ by continuously changing the threshold of the segmented image, and the larger the area covered by the curve, the better the performance of the method.

We also refer to the metric results of SCRG and BSF \cite{SCRG_1, SCRG_2}. These two metrics do not fully reflect the results of the method on target detection, because the metrics only measure groundtruth target area and cannot quantify the noise of the whole image comprehensively. Due to the characteristics of some datasets and the nature of some methods, it may cause the variance calculation of the background around the target to converge to zero and lead to the metrics statistics of infinitely large. We suppress the occurrence of metrics infinity by taking the logarithm of SCRG and BSF. The definitions of the functions for SCR, SCRG and BSF are defined as follows:
\begin{gather}
SCR = \frac{|\mu_{T} - \mu_{B}|}{\sigma}, \\
SCRG = 10\cdot\log_{10}{\frac{SCR_{out}}{SCR_{in}}}, \\
BSF = 10\cdot\log_{10}{\frac{\sigma_{in}}{\sigma_{out}}},
\end{gather}
where $|\cdot|$ denotes the absolute value of the function, $\mu_{T}$, $\mu_{B}$ denote the mean value of patch $T$ and patch $B$, respectively, and $\sigma$ denotes the standard deviation of patch $B$.

\subsection{Qualitative Evaluation}

We employed a qualitative evaluation to assess the effectiveness of our method, which focuses on detecting small moving targets in complex dynamic scenes. To this end, we present spatial, temporal, and spatial-temporal outputs, as well as 3-D feature maps, based on five IR image sequences, shown in Fig. 3. The results demonstrate the superior performance of the proposed method. Specifically, our spatial filter first extracts features that may correspond to target regions in a single frame image. Then, the temporal filter suppresses non-target regions, and the final output of a single target can be obtained through the ABS module. Moreover, the 3-D feature maps reveal that only the target patch T is enhanced, while other background and noise regions are suppressed.

We also compare the performance of seven state-of-the-art methods in single-frame detection and multi-frame detection with our proposed method on public IR image sequences, shown in Fig. 4. The first column shows the input samples of the public images, and each subsequent column displays the output feature maps of a method. In Seq. 1-4, the background is in motion due to camera movement and jittering, resulting in significant interference of the background noise on the target. Some methods fail to effectively suppress the noise and enhance the intensity of the target area while the proposed method is still able to accurately detect the target. Even in the stationary background of Seq. 5, our method still outperforms other methods in finding the target and suppressing the noise.

\subsection{Quantitative Evaluation}

In order to quantify the performance of our work, we utilized the metrics SCRG and BSF to evaluate the local enhancement of the target region by our method. Table 2 presents the performance of each method on each sequence. It can be seen from the results of the quantitative evaluation that some of the methods also suppress the surrounding background to 0 on some frames. At the same time, it is worth noting that the proposed method does not achieve the highest metrics in the target region for Seq. 4 and Seq. 5. The reason for this is the use of maximum filtering in the patch T of the spatial kernel, which enhances the area of the target, especially for very small targets. However, for larger targets, it can result in a larger area of output target enhancement, causing the edge area of the target to overlap with the background region, leading to lower metrics. Nevertheless, our method still achieved competitive scores on two metrics, as demonstrated by the results.

We also utilized the metric ROC curve to evaluate the overall target detection ability and false detection rate of the algorithm for the entire IR image sequence. As shown in Fig. 5, we compared the ROC curve of our method to other state-of-the-art methods on five public IR image sequences. The proposed method outperforms all other methods on the five IR image sequences, demonstrating its superior overall performance on the entire IR image sequence, not just the ground truth target region.  It is worth noting that our approach achieves better performance on Seq. 1-3, which have large background motion and jitter, compared to Seq. 4 and Seq. 5, where the background motion is slower or static. Despite this challenge, our method still proves effective and provides a significant improvement over existing methods.

\subsection{Ablation Study} 

We further conducted ablation experiments on our pixel-level ABS module to demonstrate the effectiveness of this strategy. It is worth noting that we propose the method mainly for the large amount of background noise caused by the synchronized motion of the background due to camera motion. Therefore when the background motion is smooth and slow, the motion of the background does not cause a great impact on the algorithm and our proposed ABS module may not play a significant boost, while when the background is static, the module is ineffective.  Fig. 6 shows the impact of the ABS module on the detection effect. The results indicate that for Seq. 1-4, the proposed method effectively improves target detection and background suppression in the presence of background motion. However, for Seq. 5, the use of the ABS module does not significantly affect detection results, either positively or negatively. As the ABS module can be inserted into any method, it can be omitted to enhance the speed of an algorithm when a priori conditions indicate that the background will not move. 

\subsection{Discussion} 

Our proposed method provides a solution for detecting weak IR small moving targets in the presence of background motion, with a focus on small point-like targets. Therefore, we perform target feature enhancement and expansion for spatial filtering. In the temporal domain, we use spatial features instead of multi-frame images for filtering and use a pixel-level ABS module to suppress the effect of motion background on target detection. Each module of our method can be integrated into other approaches, resulting in competitive performance in scenarios with scene-specific effects, although it may not achieve the best metric performance for larger volume targets or non-motion background image sequences. Our method exhibits good generalization ability.

\section{Conclusion}

In this paper, we proposed a novel method for detecting small moving targets in IR image sequences, called the spatial-temporal local feature difference method with adaptive background suppression. The method computes two feature maps using spatial and temporal filters respectively and obtains the final output feature maps after multiplying the corresponding elements by pixel-level ABS module. The experimental results demonstrate that our method is effective, accurate, and meaningful in practical application scenarios for small target detection in various types of environments.

%%%%%%%%% REFERENCES
{\small
\bibliographystyle{unsrt}
\bibliography{egbib}

\begin{thebibliography}{10}

\bibitem{detect_move_survey}
Xingchen Zhang, Ping Ye, Henry Leung, Ke~Gong, and Gang Xiao.
\newblock Object fusion tracking based on visible and infrared images: A
  comprehensive review.
\newblock {\em Information Fusion}, 63:166--187, 2020.

\bibitem{STTM-SFR}
Dongdong Pang, Pengge Ma, Tao Shan, Wei Li, Ran Tao, Yueran Ma, and Tianrun
  Wang.
\newblock Sttm-sfr: Spatial--temporal tensor modeling with saliency filter
  regularization for infrared small target detection.
\newblock {\em IEEE Transactions on Geoscience and Remote Sensing}, 60:1--18,
  2022.

\bibitem{deep_learning1}
Huan Wang, Luping Zhou, and Lei Wang.
\newblock Miss detection vs. false alarm: Adversarial learning for small object
  segmentation in infrared images.
\newblock In {\em Proceedings of the IEEE/CVF International Conference on
  Computer Vision}, pages 8509--8518, 2019.

\bibitem{deep_learning2}
Yimian Dai, Yiquan Wu, Fei Zhou, and Kobus Barnard.
\newblock Attentional local contrast networks for infrared small target
  detection.
\newblock {\em IEEE Transactions on Geoscience and Remote Sensing},
  59(11):9813--9824, 2021.

\bibitem{single_frame_survey}
Mingjing Zhao, Wei Li, Lu~Li, Jin Hu, Pengge Ma, and Ran Tao.
\newblock Single-frame infrared small-target detection: A survey.
\newblock {\em IEEE Geoscience and Remote Sensing Magazine}, 10(2):87--119,
  2022.

\bibitem{frame_difference}
Sandeep~Singh Sengar and Susanta Mukhopadhyay.
\newblock Moving object detection based on frame difference and w4.
\newblock {\em Signal, Image and Video Processing}, 11:1357--1364, 2017.

\bibitem{background_subtraction}
Sandeep~Singh Sengar and Susanta Mukhopadhyay.
\newblock Moving object detection using statistical background subtraction in
  wavelet compressed domain.
\newblock {\em Multimedia Tools and Applications}, 79(9-10):5919--5940, 2020.

\bibitem{LCM}
CL~Philip Chen, Hong Li, Yantao Wei, Tian Xia, and Yuan~Yan Tang.
\newblock A local contrast method for small infrared target detection.
\newblock {\em IEEE transactions on geoscience and remote sensing},
  52(1):574--581, 2013.

\bibitem{RLCM}
Jinhui Han, Kun Liang, Bo~Zhou, Xinying Zhu, Jie Zhao, and Linlin Zhao.
\newblock Infrared small target detection utilizing the multiscale relative
  local contrast measure.
\newblock {\em IEEE Geoscience and Remote Sensing Letters}, 15(4):612--616,
  2018.

\bibitem{MPCM}
Yantao Wei, Xinge You, and Hong Li.
\newblock Multiscale patch-based contrast measure for small infrared target
  detection.
\newblock {\em Pattern Recognition}, 58:216--226, 2016.

\bibitem{DNGM}
Lang Wu, Yong Ma, Fan Fan, Minghui Wu, and Jun Huang.
\newblock A double-neighborhood gradient method for infrared small target
  detection.
\newblock {\em IEEE Geoscience and Remote Sensing Letters}, 18(8):1476--1480,
  2020.

\bibitem{STLCF}
Lizhen Deng, Hu~Zhu, Chao Tao, and Yantao Wei.
\newblock Infrared moving point target detection based on spatial--temporal
  local contrast filter.
\newblock {\em Infrared Physics \& Technology}, 76:168--173, 2016.

\bibitem{STLDM}
Peng Du and Askar Hamdulla.
\newblock Infrared moving small-target detection using spatial--temporal local
  difference measure.
\newblock {\em IEEE Geoscience and Remote Sensing Letters}, 17(10):1817--1821,
  2019.

\bibitem{NSTSM}
Dongdong Pang, Tao Shan, Pengge Ma, Wei Li, Shengheng Liu, and Ran Tao.
\newblock A novel spatiotemporal saliency method for low-altitude slow small
  infrared target detection.
\newblock {\em IEEE Geoscience and Remote Sensing Letters}, 19:1--5, 2021.

\bibitem{STVDM}
Yunsheng Zhang, Kaijun Leng, and Kyoung-Su Park.
\newblock Infrared detection of small moving target using spatial--temporal
  local vector difference measure.
\newblock {\em IEEE Geoscience and Remote Sensing Letters}, 19:1--5, 2022.

\bibitem{data_22scene}
Bingwei Hui, Zhiyong Song, Hongqi Fan, P~Zhong, W~Hu, X~Zhang, J~Lin, H~Su,
  W~Jin, Y~Zhang, et~al.
\newblock A dataset for infrared image dim-small aircraft target detection and
  tracking under ground/air background.
\newblock {\em Sci. Data Bank}, 5:12, 2019.

\bibitem{ROC}
Jesse Davis and Mark Goadrich.
\newblock The relationship between precision-recall and roc curves.
\newblock In {\em Proceedings of the 23rd international conference on Machine
  learning}, pages 233--240, 2006.

\bibitem{SCRG_1}
Yimian Dai and Yiquan Wu.
\newblock Reweighted infrared patch-tensor model with both nonlocal and local
  priors for single-frame small target detection.
\newblock {\em IEEE journal of selected topics in applied earth observations
  and remote sensing}, 10(8):3752--3767, 2017.

\bibitem{SCRG_2}
Yuxin Hu, Yapeng Ma, Zongxu Pan, and Yuhan Liu.
\newblock Infrared dim and small target detection from complex scenes via
  multi-frame spatial--temporal patch-tensor model.
\newblock {\em Remote Sensing}, 14(9):2234, 2022.

\end{thebibliography}
}

% \end{sloppypar}
\end{document}